\documentclass{article}

\usepackage{arxiv}
\usepackage[numbers]{natbib}
\usepackage[utf8]{inputenc} 
\usepackage[T1]{fontenc}    
\usepackage{hyperref}       
\usepackage{url}            
\usepackage{booktabs}       
\usepackage{amsfonts}       
\usepackage{nicefrac}       
\usepackage{microtype}      
\usepackage{lipsum}		
\usepackage{graphicx}
\usepackage{doi}
\usepackage{amsmath}

\title{Vision Eagle Attention: A New Lens for Advancing Image Classification}

\author{ \href{https://orcid.org/0009-0007-0376-1760}{\includegraphics[scale=0.06]{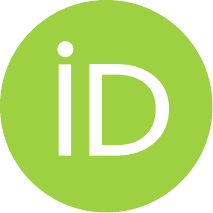}\hspace{1mm}Mahmudul Hasan}\thanks{\url{https://www.linkedin.com/in/mahmudulhasan11085/}} \\
	Department of Electrical and Electronic Engineering\\
	Bangladesh University of Engineering and Technology (BUET)\\
	Dhaka, Bangladesh 1205 \\
	\texttt{1806157@eee.buet.ac.bd} \\
}

\renewcommand{\shorttitle}{Vision Eagle Attention: A New Lens for Advancing Image Classification}

\hypersetup{
pdftitle={Vision Eagle Attention: A new era of Attention mechanism in classification},
pdfsubject={q-bio.NC, q-bio.QM},
pdfauthor={Mahmudul Hasan},
pdfkeywords={Vision Eagle Attention, Spatial Attention, ResNet-18, Image Classification, Deep Learning, FashionMNIST, OracleMNIST},
}

\begin{document}
\maketitle

\begin{abstract}
In computer vision tasks, the ability to focus on relevant regions within an image is crucial for improving model performance, particularly when key features are small, subtle, or spatially dispersed. Convolutional neural networks (CNNs) typically treat all regions of an image equally, which can lead to inefficient feature extraction. To address this challenge, I have introduced Vision Eagle Attention, a novel attention mechanism that enhances visual feature extraction using convolutional spatial attention. The model applies convolution to capture local spatial features and generates an attention map that selectively emphasizes the most informative regions of the image. This attention mechanism enables the model to focus on discriminative features while suppressing irrelevant background information. I have integrated Vision Eagle Attention into a lightweight ResNet-18 architecture, demonstrating that this combination results in an efficient and powerful model. I have evaluated the performance of the proposed model on three widely used benchmark datasets: FashionMNIST, Intel Image Classification, and OracleMNIST, with a primary focus on image classification. Experimental results show that the proposed approach improves classification accuracy. Additionally, this method has the potential to be extended to other vision tasks, such as object detection, segmentation, and visual tracking, offering a computationally efficient solution for a wide range of vision-based applications. Code is available at: \url{https://github.com/MahmudulHasan11085/Vision-Eagle-Attention.git}

\end{abstract}


\section{Introduction}
Computer vision tasks, such as image classification, object recognition, and segmentation, have made significant progress due to the development of deep learning techniques, particularly convolutional neural networks (CNNs) \citep{lecun2015deep}. CNNs have become the cornerstone of modern vision systems, as they are capable of automatically learning hierarchical features from raw image data \citep{krizhevsky2012imagenet}. However, traditional CNN architectures often treat all regions of an image equally, which can lead to suboptimal performance, especially when important features are small, spatially dispersed, or subtle \citep{hu2018squeeze}. In such scenarios, an attention mechanism that selectively focuses on discriminative regions of an image can significantly enhance feature extraction and model performance \citep{vaswani2017attention}.

Attention mechanisms have gained considerable attention in recent years due to their ability to allow models to focus on relevant parts of the input while suppressing irrelevant background information \citep{woo2018cbam}. This selective focus has been shown to improve accuracy and interpretability in a wide range of computer vision tasks \citep{zhang2019dual}. Specifically, spatial attention mechanisms focus on important spatial locations in the image, enabling models to learn more meaningful representations \cite{wang2018non}. Many of the existing attention mechanisms, however, rely on complex multi-stage processes, which can be computationally expensive and difficult to implement in lightweight models \citep{xie2017deep}.

\section{Related works}
The application of attention mechanisms in computer vision has gained significant attention due to their ability to focus on important regions within an image, improving both model performance and interpretability. The most well-known attention model is the Self-Attention mechanism \citep{vaswani2017attention}. This model computes pairwise attention scores between all pixels in the image, enabling the network to capture long-range dependencies and focus on globally relevant features. While effective, the self-attention mechanism can be computationally expensive, particularly for high-resolution images, due to its quadratic complexity in terms of image size.

To address this limitation, several approaches have been proposed to reduce the computational overhead. Non-Local Neural Networks \citep{wang2018non} use a simpler form of attention by aggregating features from all locations in the image based on their relevance. This method has been widely applied in various tasks such as object recognition and segmentation. Similarly, Squeeze-and-Excitation Networks (SENet) \citep{hu2018squeeze} introduced an efficient channel-wise attention mechanism that enhances the representational power of CNNs by adaptively recalibrating channel-wise feature responses. These methods have shown significant improvements in performance across various vision tasks, particularly in terms of classification accuracy.

Another important class of attention mechanisms is spatial attention, which focuses on important spatial regions rather than the entire image. The Convolutional Block Attention Module (CBAM) \citep{woo2018cbam} is one such example, where attention maps are learned to capture both channel-wise and spatial-wise features. The CBAM model improves the feature representation by emphasizing the most informative spatial locations in an image. 

Moreover, the Efficient Channel Attention (ECA) mechanism \citep{cao2019eca} was introduced to improve the efficiency of channel attention. ECA eliminates the need for dimensionality reduction in traditional channel attention mechanisms and instead uses local cross-channel interactions to capture channel-wise dependencies. This results in a lightweight and computationally efficient model that performs well in a variety of vision tasks, especially in resource-constrained environments.

My work builds upon the idea of spatial attention, but with a focus on creating a lightweight solution. The key contribution of my paper is the development of a convolutional-based spatial attention mechanism named Vision Eagle Attention for image classification tasks. My proposed method is evaluated on three different datasets to demonstrate its performance.

\section{Materials and Methodology}

\subsection{Datasets}
In this study, three benchmark datasets are used to evaluate the performance of the proposed Vision Eagle Attention mechanism for image classification tasks. These datasets are widely recognized in the computer vision community and provide a diverse set of challenges in terms of image complexity, object types, and dataset size.

\subsubsection{FashionMNIST}
The FashionMNIST dataset \citep{fashion} consists of 60,000 grayscale images of 10 fashion categories, each with a size of 28x28 pixels. The dataset includes clothing items such as t-shirts, trousers, dresses, and shoes, and is often used as a benchmark for image classification tasks. FashionMNIST is a relatively simple dataset compared to natural image datasets, making it an ideal choice for evaluating lightweight models and attention mechanisms. Each image is labeled with one of the 10 categories, and the task is to classify these images into their respective classes.
\subsubsection{Intel Image Classification}
The Intel Image Classification dataset \citep{intel_image_dataset} is a collection of 25,000 images belonging to six different categories: buildings, forest, glacier, mountain, sea, and street. The images are collected from various real-world environments and have a resolution of 150x150 pixels. This dataset provides a more challenging task due to the complex nature of the images, which contain multiple objects or scenes in different orientations and lighting conditions. It is used to evaluate the performance of models in more realistic image classification scenarios.
\subsubsection{OracleMNIST}
The OracleMNIST dataset \citep{oracle} consists of 28x28 grayscale images representing 30,222 ancient characters, categorized into 10 classes. The dataset contains a training set of 27,222 images and a test set of 300 images per class. OracleMNIST shares the same data format as the original MNIST dataset \citep{mnist}, making it compatible with existing classifiers and systems. However, it introduces unique challenges compared to MNIST, primarily due to the extreme noise and distortion in the images caused by over three thousand years of burial and aging. Additionally, the dataset contains a wide variation in the writing styles of ancient Chinese characters, making the classification task more complex. These characteristics make OracleMNIST a more difficult and realistic benchmark for pattern classification, particularly in scenarios involving image noise and distortion. 

\subsection{Vision Eagle Attention model}

The Vision Eagle Attention (VEA) model enhances the feature extraction capabilities of a ResNet-18 \citep{resnet18} backbone by integrating attention mechanisms at multiple stages in the network.
\begin{figure}
    \centering
    \includegraphics[width=1\linewidth,trim={10cm 10cm 10cm 10cm},clip]{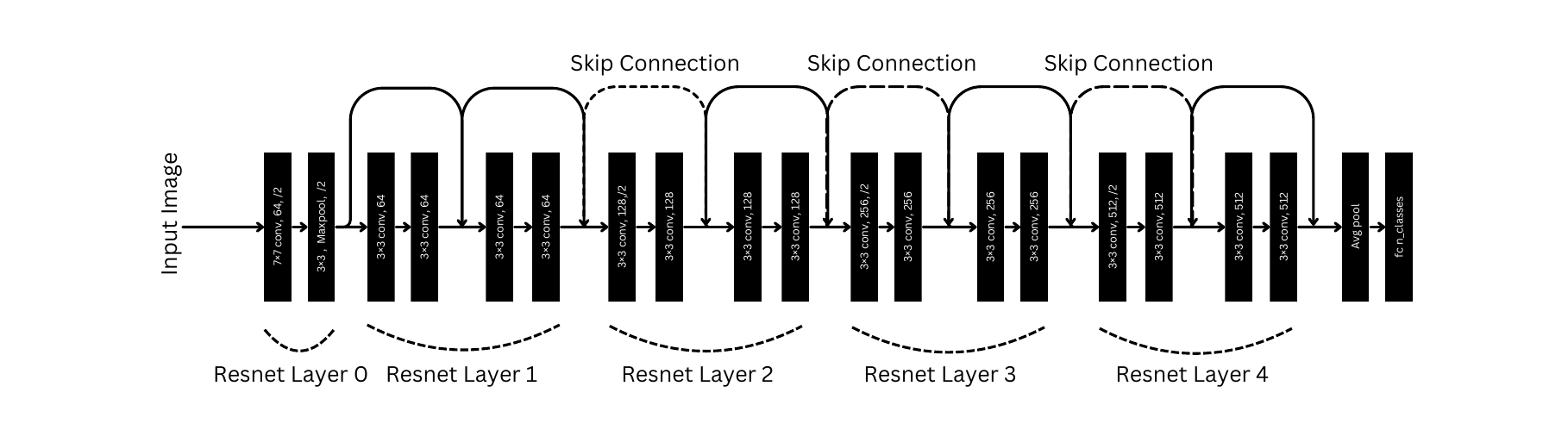}
    \caption{Architecture of ResNet-18 model.}
    \label{fig:Resnet18}
\end{figure}
The VEA model architecture consists of three Vision Eagle Attention blocks, each inserted after specific layers in the ResNet backbone. These blocks refine the feature maps progressively, allowing the network to capture more informative representations at each stage. 
\begin{figure}
    \centering
    \includegraphics[width=1\linewidth,trim={15cm 10cm 15cm 10cm},clip]{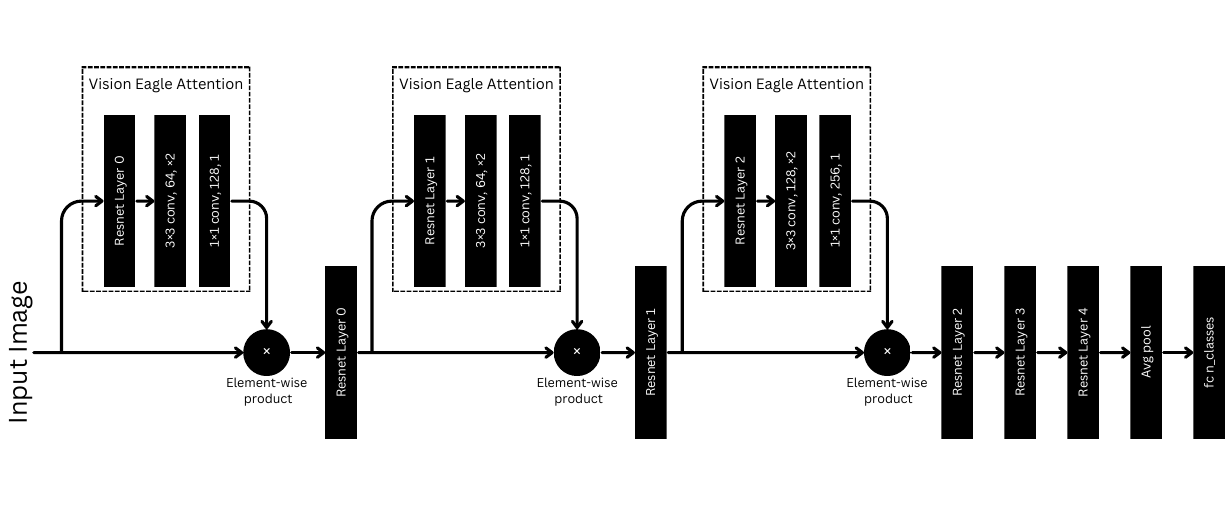}
    \caption{Architecture of Vision Eagle Attention (VEA) model.}
    \label{fig:VEA}
\end{figure}
The structure is as follows:
\begin{itemize}

    \item Vision Eagle Attention Block 1: 
    Positioned after ResNet Layer 0, this block applies a \(3 \times 3\) convolutional layer with 64 filters, followed by a \(1 \times 1\) convolution with 128 filters. The attention-modulated output is then combined with the output of ResNet Layer 0 through an element-wise product, allowing the model to enhance important features from the input image.
    \item Vision Eagle Attention Block 2: 
    Inserted after ResNet Layer 1, this block similarly uses a \(3 \times 3\) convolution with 64 filters and a \(1 \times 1\) convolution with 128 filters. Again, the output of this attention block is element-wise multiplied with the ResNet Layer 1 output to incorporate learned attention weights into the feature map.
    \item Vision Eagle Attention Block 3:
    Following ResNet Layer 2, this block uses a \(3 \times 3\) convolution with 128 filters and a \(1 \times 1\) convolution with 256 filters. The attention-modulated features are combined with ResNet Layer 2’s output through an element-wise product, guiding the model’s attention towards key image regions.
\end{itemize}
After the three VEA blocks, the feature map progresses through the deeper ResNet layers (Layers 2, 3, and 4) for further processing. The refined feature map is then passed to an average pooling layer and subsequently to a fully connected (fc) layer, which outputs the class predictions.

This multi-stage attention mechanism allows the model to adaptively emphasize important features at different depths, leading to a more focused and effective feature representation across the network.
\subsection{Implementation details and data preprocessing}
I have implemented all neural networks in this work using the PyTorch deep learning library. The experiments to evaluate the proposed framework were conducted on an NVIDIA T4 GPU, with a training batch size of 64. I have trained the model using the Stochastic Gradient Descent (SGD) optimizer, setting an initial learning rate of 0.1 for the Intel Image Classification dataset \citep{intel_image_dataset} and 0.01 for the FashionMNIST \citep{fashion} and OracleMNIST \citep{oracle} datasets. Training was conducted over 100 epochs. To optimize the learning process, I have employed a learning rate scheduler that adjusted the learning rate every 4 epochs by multiplying it by 0.5, allowing for smoother convergence and better model performance. I have also converted grayscale medical images into three-channel RGB images and resized to 100$\times$100 pixels for Intel Image Classification and OracleMNIST datasets and resized to 28$\times$28 pixels for the FashionMNIST dataset. 

\subsection{Evaluation metrics}
The following evaluation metrics are  used to assess the performance of classification model:

\begin{align}
&Accuracy = \frac{T_p + T_n}{T_p + T_n + F_p + F_n} \label{eq:11}\\
&Precision = \frac{T_p}{T_p + F_p}\label{eq:2}\\
&Sensitivity = \frac{T_p}{T_p + F_n}\label{eq:3}\\
&Specificity = \frac{T_n}{F_p + T_n} \label{eq:4}\\
&F1 = \frac{2 \cdot Precision \cdot Sensitivity}{Precision + Sensitivity} \label{eq:5}\\
&MCC = \frac{T_p \cdot T_n - F_p \cdot F_n}{\sqrt{(T_p + F_p)(T_p + F_n)(T_n + F_p)(T_n + F_n)}} \label{eq:6}
\end{align}
where \(T_p\), \(T_n\), \(F_p\), and \(F_n\) represent the counts of true positives, true negatives, false positives, and false negatives, respectively. These metrics evaluate the model's ability to classify positive and negative instances, balancing precision, recall, and overall performance.

\section{Experimental Results}

In this section, I have presented the experimental results obtained from the evaluation of two models: ResNet-18 \citep{resnet18} and Vision Eagle Attention Model, on three different datasets: FashionMNIST \citep{fashion}, Intel Image Classification \citep{intel_image_dataset}, and OracleMNIST \citep{oracle}. These datasets vary in terms of the number of classes, which are 10 for FashionMNIST and OracleMNIST, and 6 for Intel Image Classification. The results are evaluated using various metrics, including accuracy, precision, sensitivity, specificity, F1 score, and MCC.

\subsection{Comparison of Evaluation Metrics for ResNet-18 and Vision Eagle Attention Model}

Table \ref{tab:metrics_comparison} shows a comparison of the evaluation metrics for both models across the three datasets.

\begin{table}[h!]
\centering
\caption{Comparison of Evaluation Metrics for ResNet-18 and Vision Eagle Attention Model.}
\resizebox{\textwidth}{!}{%
\begin{tabular}{|c|c|c|c|c|c|c|c|}
\hline
\textbf{Model} & \textbf{Dataset} & \textbf{Accuracy} & \textbf{Precision} & \textbf{Sensitivity} & \textbf{Specificity} & \textbf{F1 Score} & \textbf{MCC} \\ \hline
ResNet-18 & FashionMNIST & 0.9228 & 0.9224 & 0.9228 & 0.9914 & 0.9224 & 0.9140 \\ \hline
Vision Eagle Attention + ResNet-18 & FashionMNIST & \textbf{0.9330} & \textbf{0.9327} & \textbf{0.9330} & \textbf{0.9926} & \textbf{0.9328} & \textbf{0.9254} \\ \hline
ResNet-18 & Intel Image Classification &0.9093 & 0.9117 & 0.9117 & 0.9818 & 0.9112 & 0.8933 \\ \hline
Vision Eagle Attention + ResNet-18 & Intel Image Classification &\textbf{0.9243} &\textbf{0.9260} &\textbf{0.9263} &\textbf{0.9848} &\textbf{0.9261} &\textbf{0.9110} \\ \hline
ResNet-18 & OracleMNIST &0.9677 & 0.9678 & 0.9677 & 0.9964 & 0.9677 & 0.9641 \\ \hline
Vision Eagle Attention + ResNet-18 & OracleMNIST &\textbf{0.9720} & \textbf{0.9720} & \textbf{0.9720} & \textbf{0.9969} & \textbf{0.9720} & \textbf{0.9689} \\ \hline
\end{tabular}%
}

\label{tab:metrics_comparison}
\end{table}

From Table \ref{tab:metrics_comparison}, I can observe that the Vision Eagle Attention model generally outperforms ResNet-18 \citep{resnet18} across all datasets, achieving higher accuracy, precision, sensitivity, specificity, F1 score, and MCC, especially on Intel Image Classification dataset.

\subsection{Confusion Matrix}

The confusion matrix for the Vision Eagle Attention model on OracleMNIST \citep{oracle} is shown in Table \ref{tab:confusion_matrix_oraclemnist}. This matrix provides a breakdown of the model's performance in terms of true positives, true negatives, false positives, and false negatives.

\begin{table}[h!]
\centering
\caption{Confusion Matrix for Vision Eagle Attention Model on OracleMNIST.}
\label{tab:confusion_matrix_oraclemnist}
\begin{tabular}{|c|c|c|c|c|c|c|c|c|c|c|}
\hline
 & \textbf{big} & \textbf{sun} & \textbf{moon} & \textbf{cattle} & \textbf{next} & \textbf{field} & \textbf{not} & \textbf{arrow} & \textbf{$\textbf{time}^{1}$} & \textbf{wood} \\ \hline
\textbf{big} & 293 & 0 & 1 & 0 & 0 & 0 & 0 & 4 & 1 & 1 \\ \hline
\textbf{sun} & 0 & 296 & 1 & 0 & 0 & 3 & 0 & 0 & 0 & 0 \\ \hline
\textbf{moon} & 1 & 1 & 295 & 0 & 0 & 0 & 2 & 1 & 0 & 0 \\ \hline
\textbf{cattle} & 0 & 0 & 0 & 285 & 4 & 0 & 1 & 0 & 5 & 5 \\ \hline
\textbf{next} & 0 & 0 & 2 & 2 & 293 & 2 & 0 & 0 & 1 & 0 \\ \hline
\textbf{field} & 0 & 3 & 0 & 0 & 1 & 296 & 0 & 0 & 0 & 0 \\ \hline
\textbf{not} & 0 & 0 & 2 & 0 & 0 & 0 & 294 & 1 & 3 & 0 \\ \hline
\textbf{arrow} & 6 & 0 & 1 & 0 & 3 & 0 & 2 & 286 & 0 & 2 \\ \hline
\textbf{$\textbf{time}^{1}$} & 0 & 2 & 1 & 1 & 1 & 0 & 0 & 1 & 291 & 3 \\ \hline
\textbf{wood} & 0 & 0 & 0 & 8 & 2 & 0 & 0 & 1 & 2 & 287 \\ \hline
\end{tabular}
\end{table}

The confusion matrix provides insight into how well the Vision Eagle Attention model identifies each class in the OracleMNIST dataset. The model performs exceptionally well, with very few misclassifications.

\subsection{Training and Inference Time Comparison}

Table \ref{tab:time_comparison} compares the training and inference times of the ResNet-18 \citep{resnet18} and Vision Eagle Attention model across the three datasets.

\begin{table}[h!]
\centering
\caption{Training and Inference Time Comparison.}
\label{tab:time_comparison}
\begin{tabular}{|c|c|c|c|}
\hline
\textbf{Model} & \textbf{Dataset} & \textbf{Training Time (s)} & \textbf{Inference Time (s)} \\ \hline
ResNet-18 & FashionMNIST & 1699 & 1 \\ \hline
Vision Eagle Attention & FashionMNIST & 2341 & 1 \\ \hline
ResNet-18 & Intel Image Classification & 2242 & 4 \\ \hline
Vision Eagle Attention & Intel Image Classification & 2448 & 6 \\ \hline
ResNet-18 & OracleMNIST & 1985 & 1 \\ \hline
Vision Eagle Attention & OracleMNIST & 2654 & 1 \\ \hline
\end{tabular}
\end{table}

In terms of training and inference time, the Vision Eagle Attention model takes slightly longer to train and infer compared to ResNet-18, but it provides superior performance across all metrics.

\section{Discussion}
The experimental results demonstrate that the Vision Eagle Attention (VEA) model outperforms the baseline ResNet-18 \citep{resnet18} architecture across all evaluation metrics and datasets. This improvement can be attributed to the enhanced spatial attention mechanism introduced by the VEA blocks, which effectively prioritize important image regions while suppressing irrelevant background noise.

In the FashionMNIST dataset \citep{fashion}, the lightweight nature of the attention mechanism allows the model to achieve remarkable accuracy improvements with minimal computational overhead. The Intel Image Classification dataset \citep{intel_image_dataset}, which features more complex and diverse image content, highlights the robustness of the VEA model in handling real-world image classification challenges. The OracleMNIST dataset \citep{oracle}, characterized by high noise and ancient character complexity, showcases the VEA model's ability to learn discriminative features even under challenging conditions.

The confusion matrix for OracleMNIST further supports these findings, indicating a high level of precision and recall for most classes, with only a small number of misclassifications. The Vision Eagle Attention mechanism's ability to emphasize subtle, spatially dispersed features proves beneficial in addressing the inherent challenges posed by this dataset. While the VEA model introduces a slight increase in training and inference times, the trade-off is justified by the significant gains in accuracy, F1 score, and MCC.

\section{Conclusion}
In this paper, I have introduced Vision Eagle Attention (VEA), a novel convolutional spatial attention mechanism designed to enhance image classification performance by integrating a lightweight convolutional attention module within a ResNet-18 \citep{resnet18} architecture. Through extensive experiments on FashionMNIST \citep{fashion}, Intel Image Classification \citep{intel_image_dataset}, and OracleMNIST datasets \citep{oracle}, the VEA model demonstrated significant improvements in accuracy, precision, F1 score, and MCC over the baseline ResNet-18 model. The results validate the effectiveness of the proposed attention mechanism in prioritizing key image regions and addressing challenges such as noise, distortion, and diverse feature representation. The model's computational efficiency and adaptability to various datasets suggest its potential for broader applications, including object detection, image segmentation, and other computer vision tasks.

Future work could explore extending the Vision Eagle Attention mechanism to other backbone architectures, incorporating it into multi-modal frameworks, or applying it to additional vision domains, such as medical imaging or remote sensing. With its lightweight design and strong performance, Vision Eagle Attention offers a promising avenue for advancing attention mechanisms in computer vision.

\bibliographystyle{unsrtnat}
\bibliography{main}

\end{document}